\documentclass[conference]{ieeeconf}
\pdfoutput=1
\IEEEoverridecommandlockouts

\usepackage[backend=biber,
            hyperref=true,
            url=false,
            isbn=false,
            doi=false,
            backref=false,
            style=ieee,
            citestyle=numeric-comp,
            sorting=nyt,
            block=none]{biblatex}
        
\renewcommand{\bibfont}{\small}
\usepackage{flushend}
\usepackage{balance}
\usepackage{marginnote}
\usepackage{graphics}
\usepackage[pdftex]{graphicx}
\usepackage{wrapfig}
\DeclareGraphicsExtensions{.pdf,.png,.jpg}

\usepackage[font={small}]{caption}
\usepackage{subcaption}
\usepackage[rightcaption]{sidecap}
\usepackage{pbox}

\usepackage{mathtools}
\usepackage{amsmath, amssymb, amscd}
\usepackage{ wasysym } 
\usepackage{amsfonts}
\usepackage{mathptmx} 
\usepackage{gensymb} 
\usepackage{nicefrac}       

\DeclareMathAlphabet{\mathcal}{OMS}{lmsy}{m}{n}
\DeclareSymbolFont{largesymbols}{OMX}{cmex}{m}{n}
\usepackage{textcomp} 
\usepackage{dsfont}

\usepackage{array} 
\usepackage{tabularx}
\usepackage{multirow}
\usepackage{multicol}
\usepackage{booktabs}

\usepackage[
  separate-uncertainty = true,
  multi-part-units = repeat
]{siunitx}
\usepackage[T1]{fontenc} 
\usepackage[utf8]{inputenc}
\usepackage[english]{babel} 
\usepackage{units}
\usepackage{bm}
\usepackage{times} 
\usepackage{xspace}
\usepackage{flushend}
\usepackage{csquotes}
\usepackage{makeidx}

\let\labelindent\relax
\usepackage{enumitem}

\usepackage{soul} 
\usepackage{subfiles} 

\usepackage[protrusion=true,expansion=true]{microtype}
\usepackage[yyyymmdd]{datetime}

\date{\protect\formatdate{1}{1}{2001}}

\usepackage{url}
\makeatletter
\g@addto@macro{\UrlBreaks}{\UrlOrds}
\makeatother
\usepackage{color}
\usepackage[usenames,dvipsnames,table]{xcolor}
\usepackage[pdfborder={0 0 0.5}]{hyperref}
\hypersetup{
    colorlinks=true,
    linkcolor=black,
    citecolor=black,
    filecolor=cyan,
    urlcolor=black
}

\usepackage{soul} 

\newcommand{\tocite}[1]{%
\textcolor{red}{[cite:\ifthenelse{\equal{#1}{}}{}{#1}?]}
}

\newcommand{\ignore}[1]{}

\usepackage{algorithm}
\usepackage{algpseudocode}

\newcommand{\sysName}{MART\xspace}
\newcommand{\sysFull}{\textsc{Multi-Arm RoboTurk}\xspace}

\begin{document}

\title{\LARGE \bf Learning Multi-Arm Manipulation Through Collaborative Teleoperation}

\author{%
Albert Tung$^{*1}$,
Josiah Wong$^{*1}$,
Ajay Mandlekar$^{1}$,
Roberto Mart\'in-Mart\'in$^{1}$, 
Yuke Zhu$^{2}$,
Li Fei-Fei$^{1}$,
Silvio Savarese$^{1}$

\thanks{ $^{*}\,$These authors contributed equally.$^{1}\,$Stanford Vision \& Learning Lab, $^{2}\,$The University of Texas at Austin.
}%
}

\maketitle

\begin{abstract}
Imitation Learning (IL) is a powerful paradigm to teach robots to perform manipulation tasks by allowing them to learn from human demonstrations collected via teleoperation, but has mostly been limited to single-arm manipulation. However, many real-world tasks require multiple arms, such as lifting a heavy object or assembling a desk. Unfortunately, applying IL to multi-arm manipulation tasks has been challenging -- asking a human to control more than one robotic arm can impose significant cognitive burden and is often only possible for a maximum of two robot arms. To address these challenges, we present \sysFull (\sysName), a multi-user data collection platform that allows multiple remote users to simultaneously teleoperate a set of robotic arms and collect demonstrations for multi-arm tasks. Using \sysName, we collected demonstrations for five novel two and three-arm tasks from several geographically separated users. From our data we arrived at a critical insight: most multi-arm tasks do not require global coordination throughout its full duration, but only during specific moments. We show that learning from such data consequently presents challenges for centralized agents that directly attempt to model all robot actions simultaneously, and perform a comprehensive study of different policy architectures with varying levels of centralization on our tasks. Finally, we propose and evaluate a base-residual policy framework that allows trained policies to better adapt to the mixed coordination setting common in multi-arm manipulation, and show that a centralized policy augmented with a decentralized residual model outperforms all other models on our set of benchmark tasks. Additional results and videos at \url{https://roboturk.stanford.edu/multiarm}

\end{abstract}

\section{Introduction}
\label{sec:intro}

Imitation learning (IL) is a powerful paradigm to teach robots to perform manipulation tasks by allowing them to learn from expert demonstrations~\cite{pomerleau1989alvinn}, but IL has mostly been limited to single-arm manipulation tasks~\cite{zhang2017deep, mandlekar2020learning}.
By contrast, many real-world manipulation tasks require multiple robot arms to operate simultaneously, such as lifting a heavy object, passing an object from one arm to the other, or assembling a desk. 
However, a limited number of works~\cite{zollner2004programming, gribovskaya2008combining, silverio2015learning} have tried to apply IL techniques to multi-arm manipulation tasks, mainly due to the difficulty of collecting single-operator demonstrations within this setting. Asking a human to control more than one robotic arm simultaneously can impose significant cognitive burden~\cite{orun2019effect} and is often only possible for two robotic arms  but no more. Furthermore, such systems can require sophisticated human-control interfaces~\cite{lipton2017baxter, laghi2018shared}, such as Virtual Reality devices which are not widely available, consequently limiting the set of users that can participate in data collection.

\begin{figure}
\setlength{\fboxrule}{1pt}
\setlength{\fboxsep}{0pt}
\centering
\begin{subfigure}{0.23\textwidth}
\centering
\fbox{\includegraphics[width=\columnwidth]{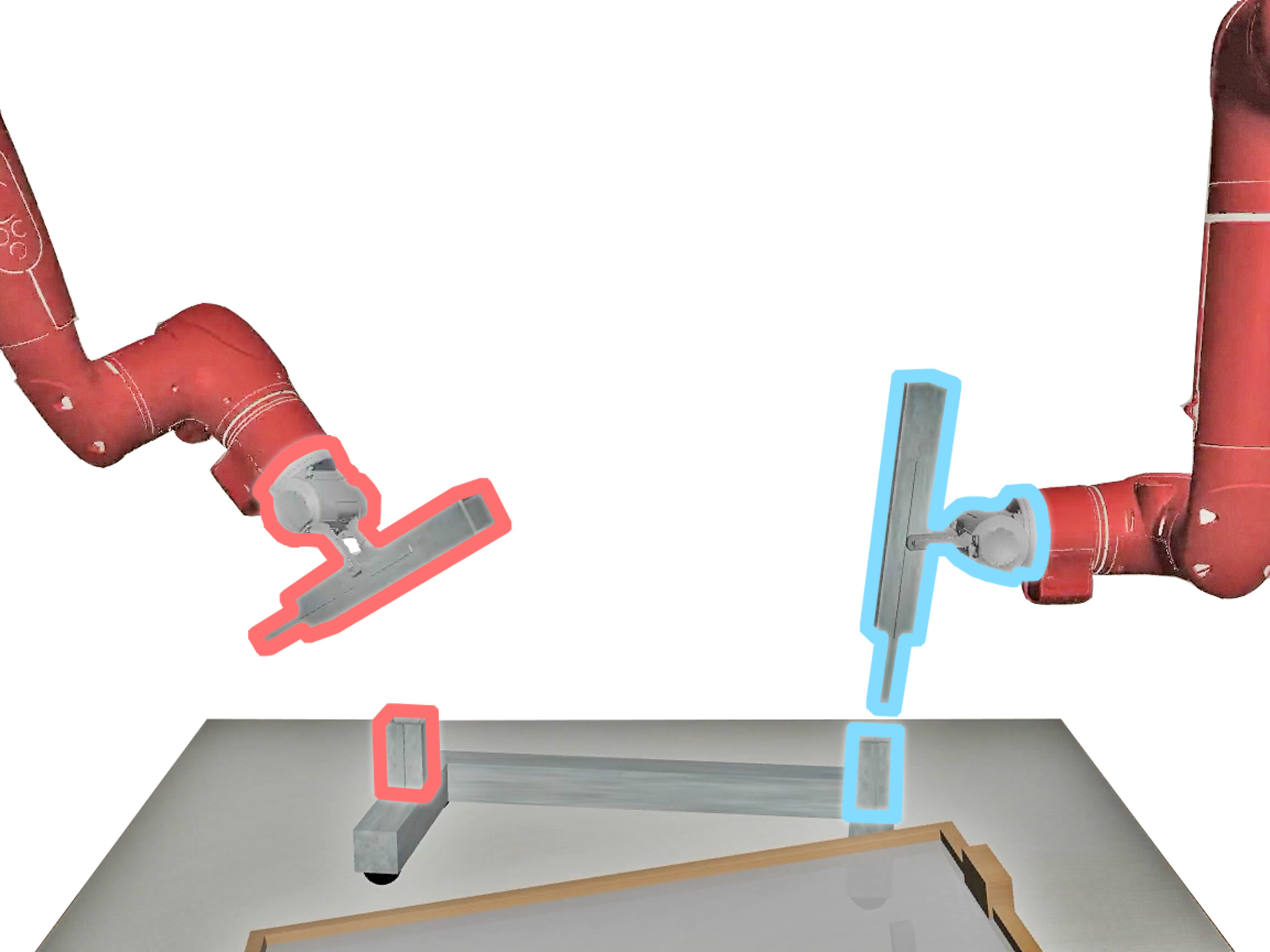}}
\end{subfigure}
\hfill
\begin{subfigure}{0.23\textwidth}
\centering
\fbox{\includegraphics[width=\columnwidth]{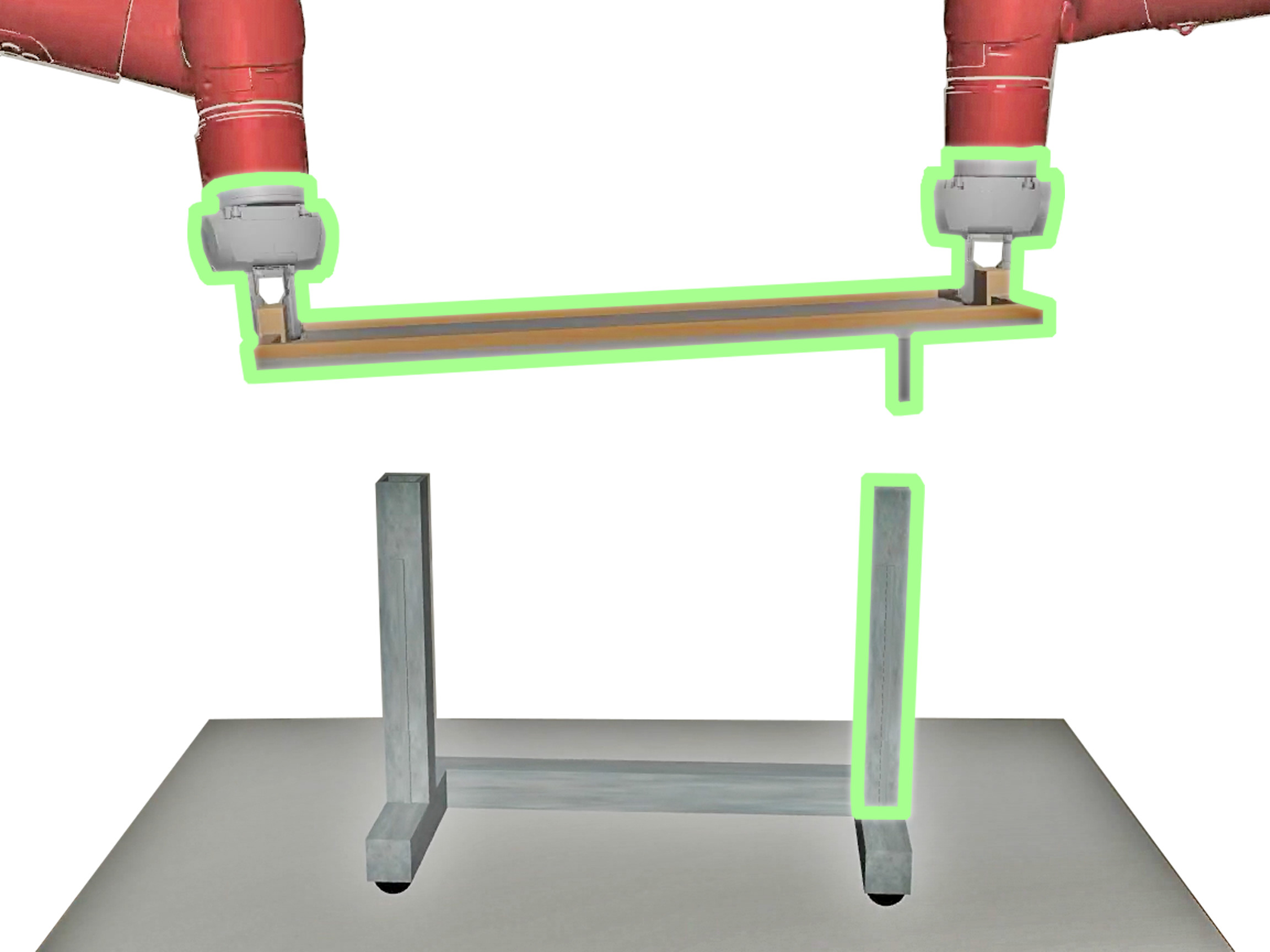}}
\end{subfigure}
\caption{\textbf{Multi-Stage Multi-Arm Manipulation with Mixed Coordination.} Table assembly is a canonical example of a multi-stage mixed coordinated task, where each arm must complete an independent, parallelized column assembly subtask \textit{(left)}, after which each arm must coordinate to lift and align the tabletop component to complete the task \textit{(right)}. We build a system that allows for remote teleoperators to collaboratively collect task demonstrations on such multi-stage multi-arm manipulation tasks.}
\label{fig:pullfig}
\vspace{-15pt}
\end{figure}

To address these limitations, we present \sysFull (\sysName), a multi-user data collection platform that allows multiple remote users to simultaneously teleoperate a set of robotic arms and collect demonstrations for multi-arm tasks. \sysName addresses the limitations of prior multi-arm systems because it frees users from cognitive burden by only having each control a single arm, allowing demonstration collection for multi-arm tasks while only requiring users to have access to a smartphone and web browser. Thus, \sysName lowers the barriers to entry for exploring the wider taxonomy of multi-arm tasks, and allowed us to collect demonstrations for five novel two-arm and three-arm tasks from users physically separated by thousands of kilometers.


After collecting and analyzing human demonstration data from these tasks, we gained the following critical insight: most multi-arm tasks do not require global coordination throughout its full duration. Consider a table assembly task (Fig~\ref{fig:pullfig}) in which each leg can be assembled independently but requires coordinated execution when aligning the tabletop. Is coordination explicitly necessary throughout?
To explore this claim, we performed extensive experiments training state-of-the-art IL variants with different levels of centralized and distributed control, representing explicit coordination and fully decoupled execution, respectively.




We \textit{a priori} expected that centralized versions should be able to coordinate actions from multiple arms the best and outperform other variants. 
However, we observed that centralized agents perform poorly across several tasks compared to distributed variants. We hypothesize this may be caused by the centralized agent ``hallucinating'' incorrect correlations between arms from the limited set of demonstrations, 
rendering the task harder than it really is.
While distributed agents do not suffer from this limitation, we observed that distributed agents can struggle to learn sections of a task where more than one arm needs to synchronize to accomplish the goal.


To address both of these issues, we propose a method for directly modeling both centralized and decoupled policies via a base-residual model trained in a two step process. Our guiding intuition is that the base policy's architecture choice can dictate the either fully coordinated or fully decoupled dominating behavior, while the residual policy can encourage the resulting composite policy to exhibit desired complementary traits. The composite policy mitigates overfitting in the centralized base policy case via a decentralized residual architecture and improves coordination in the decentralized base policy case via a centralized residual architecture . 
Our experiments demonstrate that using this augmented policy structure outperforms baselines that are fully centralized or decentralized across all of our benchmark tasks that require mixed coordination.

In summary, our contributions are as follows:
\begin{enumerate}[wide, labelwidth=!, labelindent=0pt]
    \item We present \sysFull (\sysName), a scalable multi-agent data collection system that allows us to gather demonstrations on diverse multi-arm tasks from humans remotely located via an easy and intuitive interface, lowering the barriers to entry for exploring the wider taxonomy of multi-arm tasks.
    \item We provide a set of novel realistic multi-arm benchmark tasks ranging from the fully decoupled to fully coordinated setting that allow us to analyze these emergent mixed coordination properties, including a three-arm task that, to our knowledge, is the first of its kind.
    \item We collect and evaluate human demonstrations on simulated versions of our tasks\footnote{Our system can be used ``as is'' for the collection with real-world robots, similar to Mandlekar et al.~\cite{mandlekar2019scaling}. However, due to the current COVID measures, we could not get access to our robots.} against multiple baselines, and show that fully centralized or decentralized policy models suffer during tasks requiring mixed coordination.
    \item We propose and evaluate a base-residual policy framework that allows policy models to better adapt to the mixed coordination setting, and show that policies augmented with this model are able to outperform all prior baselines across all of our tasks.
\end{enumerate}

\section{Related Work}
\label{sec:related}

\begin{figure}[t!]
    \centering
    \vspace{1mm}
     \includegraphics[width=0.9\linewidth]{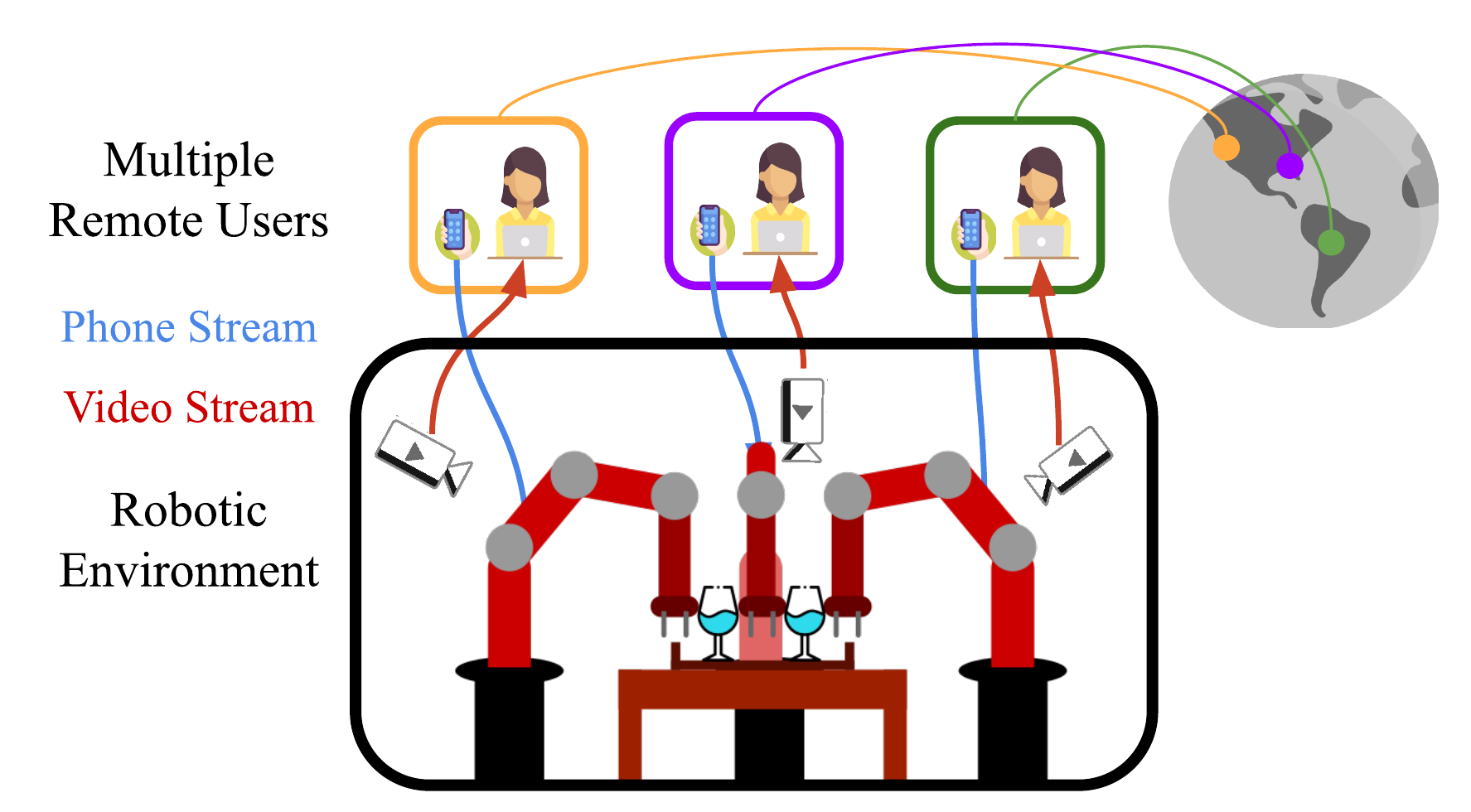}
    \caption{\textbf{Multi-Arm RoboTurk System Diagram.} Our system enables multiple remote users physically separated by thousands of kilometers to collaboratively teleoperate robot arms and collect multi-arm task demonstrations. Each operator uses their smartphone to control one robot arm and receives a video stream, tailored to a specific robot arm viewpoint, in their web browser.}
    \label{fig:system-diagram}
    \vspace{-15pt}
\end{figure}

\textbf{Multi-Agent Reinforcement Learning:} Multi-Agent Reinforcement Learning~\cite{tan1993multi, busoniu2008comprehensive} in cooperative settings has been widely studied~\cite{lowe2017multi, foerster2016learning, mataric1997reinforcement, sukhbaatar2016learning, foerster2017counterfactual, jiang2018learning}, and applied to domains such as video games~\cite{peng2017multiagent} and visual question answering~\cite{das2017learning}. Exploration in such settings can be more burdensome than in the single-agent setting due to the larger action space and dependence between agent actions.

\begin{figure*}[t]
\centering
\begin{subfigure}{0.15\textwidth}
\centering
\includegraphics[width=\columnwidth]{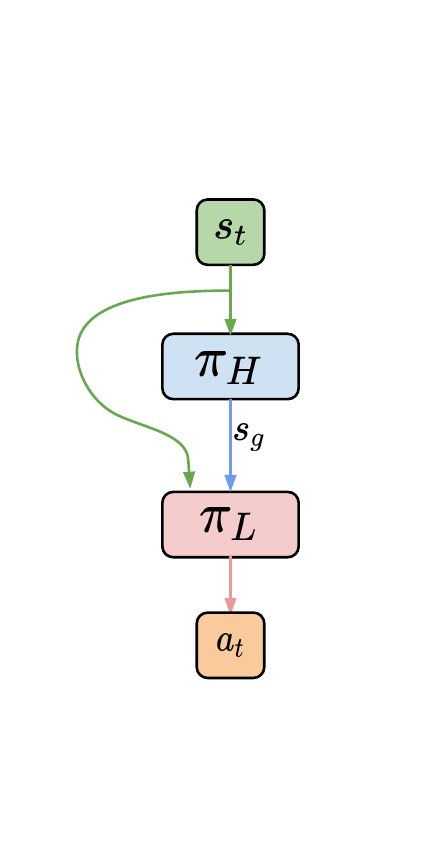}
\caption{HBC} 
\end{subfigure}
\hfill
\begin{subfigure}{0.15\textwidth}
\centering
\includegraphics[width=\columnwidth]{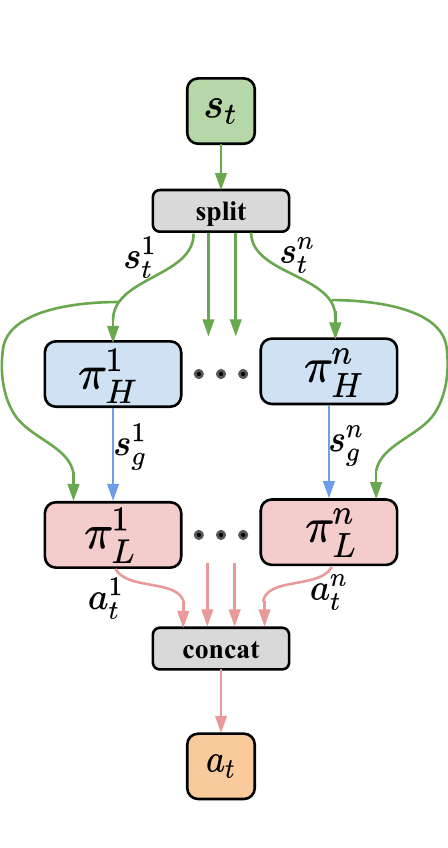}
\caption{d-HBC} 
\end{subfigure}
\hfill
\begin{subfigure}{0.15\textwidth}
\centering
\includegraphics[width=\columnwidth]{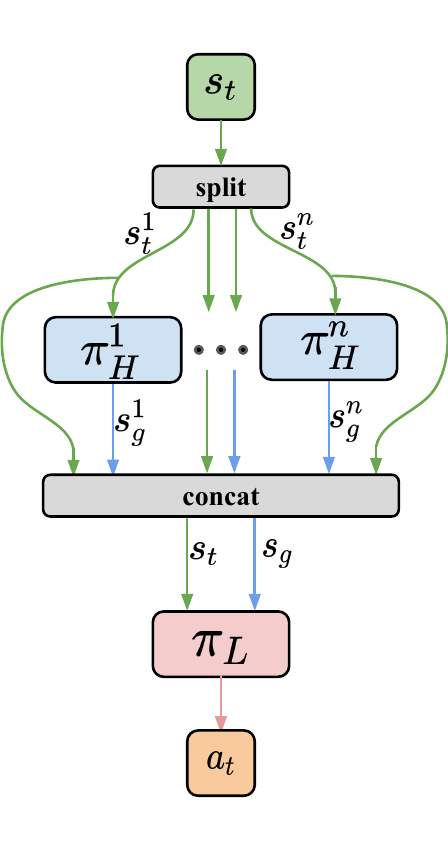}
\caption{dh-HBC} 
\end{subfigure}
\hfill
\begin{subfigure}{0.15\textwidth}
\centering
\includegraphics[width=\columnwidth]{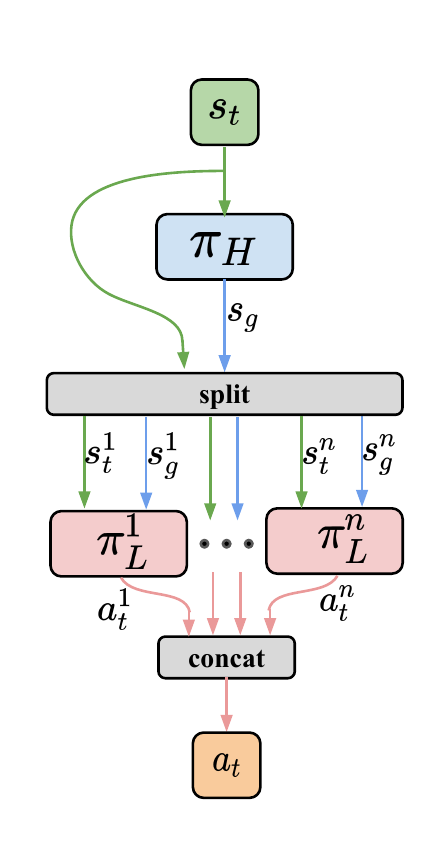}
\caption{dl-HBC} 
\end{subfigure}
\hfill
\begin{subfigure}{0.15\textwidth}
\centering
\includegraphics[width=\columnwidth]{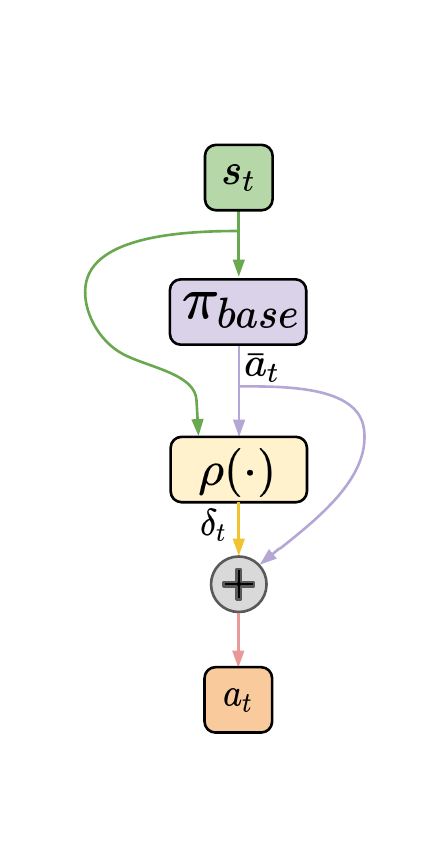}
\caption{r-HBC (ours)} 
\end{subfigure}
\hfill
\begin{subfigure}{0.15\textwidth}
\centering
\includegraphics[width=\columnwidth]{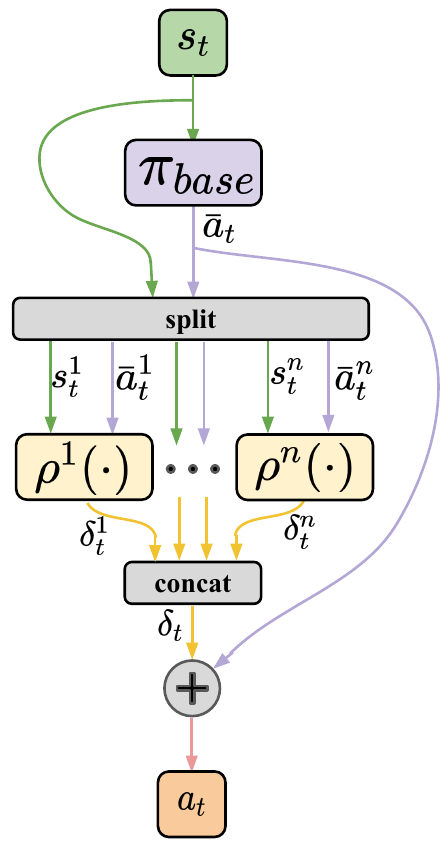}
\caption{rd-HBC (ours)} 
\end{subfigure}
\caption{\textbf{Model Architectures.} (a) Hierarchical Behavioral Cloning (HBC) where a high-level policy predicts subgoals (future observations) for all arms, and a low-level policy conditions on the subgoal. (b) Decentralized variant of HBC (d-HBC) where a separate model is trained for each arm. (c) Decentralized high-level policy with centralized low-level policy (dh-HBC). (d) Centralized high-level policy with decentralized low-level policy (dl-HBC). (e) (ours) A centralized residual network applies small perturbations to the actions from a d-HBC base policy. (f) (ours) A decentralized residual network applies small perturbations to the actions from an HBC base policy.}
\label{fig:algos}
\vspace{-15pt}
\end{figure*}

\textbf{Multi-Agent Imitation Learning:} Most work in Multi-Agent Imitation Learning~\cite{song2018multi, le2017coordinated, vsovsic2016inverse, bogert2014multi} focuses on the paradigm of Inverse Reinforcement Learning~\cite{abbeel2004apprenticeship, abbeel2011inverse}, in which multi-agent demonstrations are used to infer a reward function, and the reward function is optimized via Reinforcement Learning (RL). However, this can require extensive agent interaction due to the RL process. Chernova et al.~\cite{chernova2007multiagent} has also explored multi-agent imitation learning in an interactive setting, where humans can provide corrective actions to the agent, but the method was demonstrated on simple 2D domains. Instead, we focus on Behavioral Cloning (BC)~\cite{pomerleau1989alvinn}, a common approach for imitation learning that trains a policy from a demonstration dataset in an offline manner. 

While centralized and decentralized structures for policies and reward functions have been studied extensively in the multi-agent IRL setting~\cite{song2018multi}, they have not been explored significantly in BC settings. In general, learning from multi-arm demonstrations on manipulation tasks is unexplored.


\textbf{Bimanual Robot Manipulation:} Bimanual manipulation is a practical problem of great interest~\cite{smith2012dual}. Reinforcement Learning (RL) has been applied to bimanual manipulation tasks~\cite{kroemer2015towards, amadio2019exploiting, chitnis2020efficient, chitnis2020intrinsic}, but RL methods must deal with the increased burden of exploration due to the presence of two arms. Prior work has tried to address the exploration burden by assuming access to parametrized skills such as reaching and twisting~\cite{chitnis2020efficient}, by encouraging efficient exploration via intrinsic motivation~\cite{chitnis2020intrinsic}, and leveraging movement primitives from human demonstrations~\cite{amadio2019exploiting}. RL in this setting has mainly been limited to short-horizon single-stage tasks such as twisting a bottle cap. By contrast, in our work, by collecting human demonstrations, we are able to circumvent the exploration burden and train performant policies on challenging, multi-stage, multi-arm manipulation tasks.

Imitation Learning (IL) on bimanual tasks is less common. Some prior works~\cite{zollner2004programming, gribovskaya2008combining, silverio2015learning} have leveraged the paradigm of programming by demonstration (PbD), but these approaches often requires extensive modeling assumptions, and may not generalize well to different environment configurations.

Systems allowing for bimanual teleoperation are relatively uncommon. Laghi et al.~\cite{laghi2018shared} built a system that allows a user to simultaneously control two robot arms using special sensors that track the user's arms. Lipton et al.~\cite{lipton2017baxter} built a system that allows a remote teleoperator to control a bimanual Baxter robot using a Virtual Reality (VR) interface. Unlike \sysName, neither of these systems are suitable for multi-arm settings with more than two arms, and both rely on special purpose hardware that is not widely available, restricting the set of people that can use the system. Bimanual manipulation has also been studied in the context of assistive settings~\cite{edsinger2007two}.

\section{Preliminaries}
\label{sec:problem}

We formalize the problem of solving a robot manipulation task as an infinite-horizon discrete-time Markov Decision Process (MDP), $\mathcal{M} = (\mathcal{S}, \mathcal{A}, \mathcal{T}, R, \gamma, \rho_0)$, where $\mathcal{S}$ is the state space, $\mathcal{A}$ is the action space, $\mathcal{T}(\cdot | s, a)$, is the state transition distribution, $R(s, a, s')$ is the reward function, $\gamma \in [0, 1)$ is the discount factor, and $\rho_0(\cdot)$ is the initial state distribution. At every step, an agent observes $s_t$, uses a policy $\pi$ to choose an action, $a_t = \pi(s_t)$, and observes the next state, $s_{t+1} \sim \mathcal{T}(\cdot | s_t, a_t)$, and reward, $r_t = R(s_t, a_t, s_{t+1})$. The goal is to learn an policy $\pi$ that maximizes the expected return: $\mathbb{E}[\sum_{t=0}^{\infty} \gamma^t R(s_t, a_t, s_{t+1})]$. 

We tackle the problem of multi-robot manipulation; we assume this corresponds to a factorization of the states and actions for each robot $s = (s^1, s^2, \dots, s^n)$, $a = (a^1, a^2, \dots, a^n)$.
In this setting, we define a \textit{centralized} agent as an agent that uses the entire state, $s$, to generate an action, $a$, for all robots, and a \textit{decentralized} agent as an agent that generates each robot-specific action, $a^i$, by only using the corresponding robot observation, $s^i$. Consequently, a centralized agent uses the observation from all robot arms to jointly determine each robot's action, while a decentralized agent independently generates each robot action without considering observations from the other robot arms.

As our goal is to leverage demonstrations gathered from our novel system, we now briefly review offline imitation learning methods that can be used to learn from human demonstrations. Behavioral Cloning (BC)~\cite{pomerleau1989alvinn} is a common and simple method for learning from a set of demonstrations $\mathcal{D}$. It trains a policy $\pi_{\theta}(s)$ to learn the actions in the demonstrations with the objective: $\arg\min_{\theta} \mathbb{E}_{(s, a) \sim \mathcal{D}} ||\pi_{\theta}(s) - a||^2$. Hierarchical Behavioral Cloning (HBC) seeks to learn hierarchical policies that encourage temporal abstraction and can be a better way to learn from offline human demonstrations~\cite{mandlekar2020iris, mandlekar2020learning}. HBC consists of a low-level policy that is conditioned on future observations $s_g \in \mathcal{S}$ (termed \textit{subgoals}) and learns sequences of actions that can be used to achieve them, and a high-level policy that predicts future subgoals given a current observation. The low-level policy is a subgoal-conditioned recurrent neural network (RNN) $\pi_L(s , s_g)$ that is trained on $T$-length temporal state-action sequences to produce an action sequence $a_t, \dots, a_{t + T - 1}$, conditioned on the state sequence $s_t, \dots, s_{t + T - 1}$, and the subgoal $s_{t + T}$. The high-level policy $\pi_H(s)$ is trained to predict subgoal observations $s_{t + T}$ that are $T$ timesteps in the future from the current observation $s_t$, and is often a conditional Variational Autoencoder (cVAE)~\cite{kingma2013auto} that learns a conditional distribution $\pi_H(s_{t + T} | s_t)$~\cite{mandlekar2020iris, mandlekar2020learning}.

\section{MART: Multi-Arm RoboTurk}
\label{sec:system}

In this section, we first review the RoboTurk platform, and then show how we extended it to develop \sysName (Fig.~\ref{fig:system-diagram}).

\subsection{RoboTurk Overview}


RoboTurk~\cite{mandlekar2018roboturk, mandlekar2019scaling} is a platform that allows remote users to collect real or simulated task demonstrations through low-latency teleoperation. Users log in to a website with a real-time video stream of the robot workspace from their robot's unique vantage point, and control their robot's end effector using their smartphone as a 6-DoF motion controller. 

\begin{figure*}
\centering
\begin{subfigure}{0.19\textwidth}
\centering
\includegraphics[width=\columnwidth]{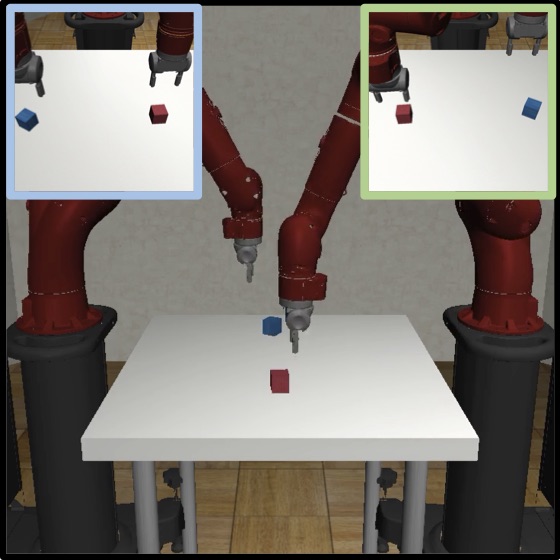}
\caption{Multi-Cube Lifting} 
\end{subfigure}
\hfill
\begin{subfigure}{0.19\textwidth}
\centering
\includegraphics[width=\columnwidth]{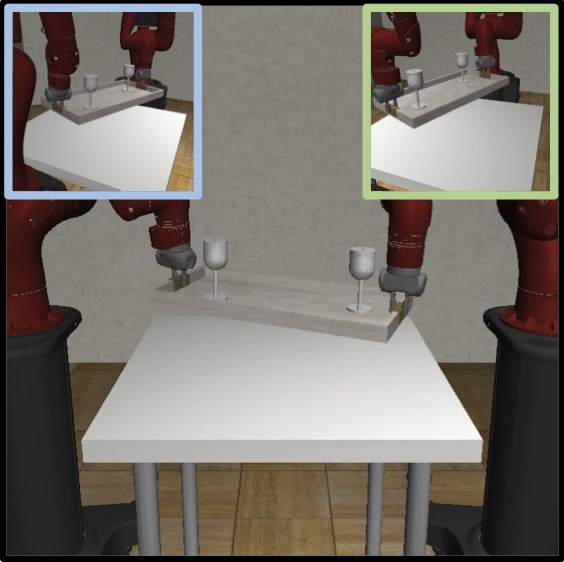}
\caption{Drink Tray Lifting} 
\end{subfigure}
\hfill
\begin{subfigure}{0.19\textwidth}
\centering
\includegraphics[width=\columnwidth]{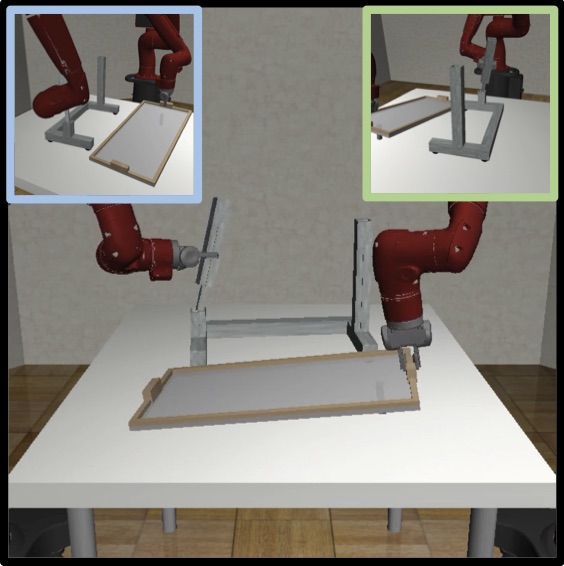}
\caption{Table Assembly} 
\end{subfigure}
\hfill
\begin{subfigure}{0.19\textwidth}
\centering
\includegraphics[width=\columnwidth]{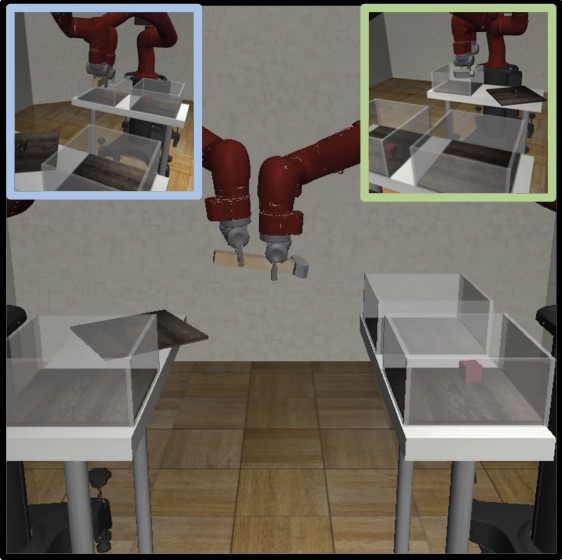}
\caption{Pick-Place Handover} 
\end{subfigure}
\hfill
\begin{subfigure}{0.19\textwidth}
\centering
\includegraphics[width=\columnwidth]{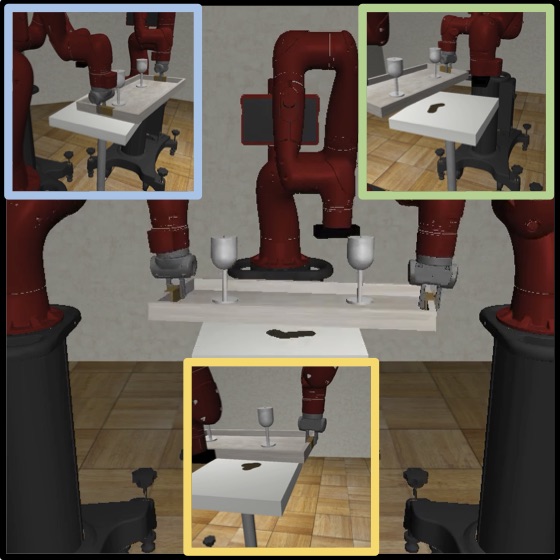}
\caption{Lifting Wiping} 
\end{subfigure}
\caption{\textbf{Tasks and Operator Viewpoints.} We present 5 challenging multi-arm manipulation tasks along with corresponding agent-specific teleoperator viewpoints. Our tasks showcase real-world scenarios requiring varying levels of coordination between agents.}
\label{fig:benchmark_tasks}
\vspace{-15pt}
\end{figure*}

To facilitate low-latency video streaming to each user's web browser, the platform leverages Web Real-Time Communication (WebRTC) to establish low-latency communication links between a user's web browser, smartphone, and the remote teleoperation server which interfaces with the robot environment. We summarize the main platform components:

\textit{Teleoperation Server:} A process dedicated to a single user that interfaces with the user endpoint and the robot. It maintains its own robot simulator instance and two WebRTC connections -- one to the user's phone, and another to the user's web browser. It uses the first connection to receive phone commands and control the robot arm and the second connection to send rendered frames of the robot workspace to the user's web browser.

\textit{User Endpoint:} The user views a video stream of the workspace in their web browser and controls the robot arm by moving their smartphone in free space. The phone pose is mapped to an end effector command.

\subsection{Extending RoboTurk for Collaborative Teleoperation}

Extending RoboTurk to incorporate multiple robotic manipulators and enable real-time user collaboration required important system design considerations (Fig.~\ref{fig:system-diagram}).

\textit{Collaborative Teleoperation:} To enable multiple users to control robot arms in the same workspace, we extended the teleoperation server to maintain multiple communication channels -- two per user, one to each user's phone and the other to each user's web browser. The server receives phone commands from each user and uses some synchronization logic to determine when to send commands to the simulated robot arms (described below). It also renders user-specific viewpoints from cameras in the workspace (see Fig.~\ref{fig:benchmark_tasks}) and sends each to the corresponding user's web browser.


\textit{Robot Command Synchronization:} To facilitate teleoperation that feels natural, we would like users to perceive that simulation is real-time (e.g. 1 second of simulation time takes 1 second). However, robot simulation is discrete-time, and requires controls for all robot arms to proceed. Unfortunately, controlling multiple arms in a single simulation from multiple phones creates a synchronization issue because of variable latency in each user's network connection. Phone commands from the different users can be received by the teleoperation server at different rates and different times. To address this issue, we wait for new phone messages to be received on all phone connections before actuating all robot arms and proceeding to the next timestep. We found this synchronization to be extremely helpful at ensuring that each user perceives simulation to run in real-time.


\section{Learning Mixed Coordination}
\label{sec:algorithm}

After collecting and analyzing demonstrations collected by \sysName, we observed that most multi-arm tasks do not require global coordination throughout their full duration, and instead only require coordination during specific subtask segments. Centralized policies that directly model the full joint state-action mapping are liable to overfit in sections that do not require coordination. To better address the problem of learning from these mixed-coordination demonstrations, we develop several variants of HBC (Fig~\ref{fig:algos}a) that combine centralized and decentralized components, as described below.
\textbf{Full Decentralization (d-HBC):} We consider per-arm policy models and partition our collected demonstrations into arm-specific observations and actions (Fig~\ref{fig:algos}b). This means training high-level policies $\pi^1_H(s^1), \dots, \pi^n_H(s^n)$ and low-level policies $\pi^1_L(s^1, s_g^1), \dots, \pi^n_L(s^n, s_g^n)$ -- one per robot arm. This architecture is fully decentralized as each set of policies generates an arm action purely from that arm's observation, disregarding other arms completely.


\textbf{Partial Decentralization (d[h/l]-HBC):} We outline a simple modification to HBC that allows for \textit{partial} decentralization. We establish two variants by factorizing either (1) the high-level policy or (2) the low-level policy to be decentralized. Notice that this is a compromise between centralized HBC, where nothing is factorized, and decentralized HBC (d-HBC), where both are factorized. In dh-HBC (Fig~\ref{fig:algos}c), the high-level is decentralized -- $n$ high-level policies produce subgoals $s_g = (s_g^1, \dots, s_g^n)$ which are fed to a centralized low-level policy $\pi_L(s, s_g)$. In dl-HBC (Fig~\ref{fig:algos}d), the high-level policy is centralized and the low-level policy is decentralized -- $n$ low-level policies produce arm actions $(a^1, \dots, a^n)$. 

\textbf{Mixed Coordination with Residual Learning (r[d]-HBC):} A more nuanced approach is to endow a pretrained policy with desired properties through a separate residual network that perturbs its action. In this way, we can choose complementary architectures that help mitigate the underlying pitfalls of the base policy architecture -- thus, if the base policy is centralized, then we provide agent-specific residual networks to reduce overfitting and encourage greater generalization. Conversely, we can provide a centralized residual network for a decentralized base policy to facilitate coordination in sections of the task that may need it. Concretely, given an action from a pretrained policy $\bar{a} = \pi(s)$, our residual network $\rho(\bar{a}, s)$ takes this action and the state as input, and outputs a small correction to the action
\begin{align}
    \label{eq:res}
    a = \bar{a} + \delta, \quad \delta = \rho(\bar{a}, s), \quad ||\delta||_2 < \epsilon, \: \text{$\epsilon$ small}
\end{align}
where we constrain the L2 norm of the perturbation to be smaller than $\epsilon$ to prevent the residual network from dominating the overall policy behavior. This results in two variants -- r-HBC (Fig~\ref{fig:algos}e), where we train a decentralized HBC base policy and then learn a centralized residual network, and rd-HBC (Fig~\ref{fig:algos}f), where we train a centralized HBC base policy and then learn a decentralized residual network.

\section{Experimental Setup}
\label{sec:exp}


In this section, we describe our benchmark multi-arm tasks, and data collection setup.


\textbf{Tasks:} All tasks were designed using MuJoCo~\cite{todorov2012mujoco} and the robosuite framework~\cite{robosuite2020} (see Fig.~\ref{fig:benchmark_tasks}). All robot arms are controlled using Operational Space Controllers~\cite{khatib1987unified}. Observations contain per-robot end-effector pose and task-specific object information. For decentralized setups, we partitioned the state space based on information relevant to each agent.

{\textit{Two Arm Multi-Cube Lifting:}} Two robot arms must lift two blocks placed on a table. This pedagogical task is fully \textit{decoupled} since each arm can lift a block independently.

{\textit{Two Arm Drink Tray Lifting:}} Two robot arms must lift and hold a tray for 1.5 seconds without tipping the drinks on the tray over. This pedagogical task represents the fully \textit{coordinated} case where each arm must consider the other's actions in order to carefully lift and stabilize the tray.

{\textit{Two Arm Assembly:}} Two robot arms must assemble a hospital bed composed of a base, two columns, and tabletop. The arms need to place the columns in the base and then coordinate to lift and align the tabletop over the columns. This task is challenging for several reasons - it is multi-stage and requires fine-grained manipulation for assembling the columns and table with varying levels of coordination over the task. The columns can be assembled independently by each arm, but the tabletop assembly requires coordination.

{\textit{Two Arm Pick-Place Handover:}} Two robot arms must work together to transfer a hammer from a closed container on a shelf to a target bin on another shelf. One robot arm must retrieve the hammer from the closed container, while the other arm must simultaneously clear the target bin by moving a cube (trash) to a nearby receptacle. Finally, one arm hands the hammer over to the other arm to place in the target bin. This task is challenging because it is multi-stage and contains subtasks that require different levels of coordination. 

{\textit{Three Arm Lift Wiping:}} A dirty tabletop must be cleaned, but has a tray of drinks on top of it. Two arms must lift and move the tray without spilling the drinks while a third arm wipes the patch of dirt on the table underneath. Solving this task requires asymmetrical coordination -- two arms must coordinate to move the tray out of the way without spilling the drinks while the third arm can operate in parallel, wiping the tabletop when the tray is cleared. 


\textbf{Data Collection:} We collect a set of experienced user demonstrations on all five novel tasks, as well as additional demonstrations on our three mixed coordination tasks from multiple user groups with varying levels of experience as part of a user study. Our user study consists of three unique user pairs for the two arm tasks, and two unique groups of three for the three arm task, with each dataset consisting of roughly 50-100 successful demonstrations.

\begin{table}
  \caption{\textbf{Marginal Success Rate Degradation vs. Subtask Type}}
  \label{table:coordination}
  \centering
  \begin{tabular}{cccc}
    \toprule
    Subtask Type &
    \begin{tabular}[c]{@{}c@{}}Three Arm\\ Lifting Wiping\end{tabular} &
    \begin{tabular}[c]{@{}c@{}}Two Arm\\ Pick-Place \\ Handover\end{tabular} &
    \begin{tabular}[c]{@{}c@{}}Two Arm\\ Table Assembly\end{tabular} \\
    \midrule
    Uncoordinated & $29.2$ & $33.8$ & $38.3$ \\
    Coordinated & $38.2$ & $12.1$ & $12.6$ \\

    \bottomrule
  \end{tabular}
\end{table}


\section{Results}
\label{sec:results}

In this section, we analyze our novel contributions, and show that (a) users can effectively coordinate using MART, and (b) our residual framework is able to outperform all other baseline models across all of our multi-arm tasks.

\subsection{System Analysis: Do operators have a hard time with coordination?}


\begin{table*}
  \caption{\textbf{Results on Experienced Operator Demonstrations}}
  \label{task_benchmark_algo_results}
  \centering
  \begin{tabular}{cccccc}
    \toprule
    Model & \begin{tabular}[c]{@{}c@{}}Two Arm\\ Multi-Cube Lifting\end{tabular} & \begin{tabular}[c]{@{}c@{}}Two Arm\\ Drink-Tray Lifting\end{tabular} & \begin{tabular}[c]{@{}c@{}}Two Arm\\ Table Assembly\end{tabular} & \begin{tabular}[c]{@{}c@{}}Two Arm\\ Pick-Place Handover\end{tabular} & \begin{tabular}[c]{@{}c@{}}Three Arm\\ Lifting Wiping\end{tabular} \\
    \midrule
    BC-RNN & $34.1\pm6.4$ & $50.0\pm12.5$ & $0.0\pm0.0$ & $2.0\pm2.7$ & $56.0\pm6.5$ \\
    HBC & $38.5\pm3.8$ & $70.7\pm3.3$ & $5.2\pm3.0$ & $16.0\pm3.9$ & $83.7\pm6.5$ \\
    d-HBC & $84.5\pm 2.5$ & $72.6\pm11.4$ & $4.9 \pm 3.1$ & $24.4 \pm 6.4$ & $50.0\pm9.3$ \\
    dh-HBC & $41.3\pm4.5$ & $55.3\pm8.8$ & $1.3\pm1.6$ & $15.3\pm4.5$ & $62.0\pm8.8$ \\
    dl-HBC & $58.0\pm10.0$ & $55.3\pm6.5$ & $5.3\pm4.0$ & $10.7\pm4.4$ & $55.3\pm18.2$ \\
    r-HBC (ours) & $58.7\pm4.5$ & $75.3\pm5.8$ & $\mathbf{10.0\pm2.1}$ & $\mathbf{31.3\pm4.0}$ & $\mathbf{94.0\pm2.5}$ \\
    rd-HBC (ours) & $\mathbf{89.3\pm3.3}$ & $\mathbf{86.7\pm3.0}$ & $5.3\pm2.7$ & $26.7\pm4.7$ & $58.6\pm1.6$ \\
    \bottomrule
  \end{tabular}
\vspace{-5pt}
\end{table*}
\begingroup
\setlength{\tabcolsep}{5pt} 
\begin{table}
  \caption{\textbf{Results on User Study Demonstrations}}
  \label{task_user_study_algo_results}
  \centering
  \begin{tabular}{cccc}
    \toprule
    Model & \begin{tabular}[c]{@{}c@{}}Two Arm\\ Table-Assembly\end{tabular} & \begin{tabular}[c]{@{}c@{}}Two Arm\\ Pick-Place \\ Handover\end{tabular} & \begin{tabular}[c]{@{}c@{}}Three Arm\\ Lifting Wiping\end{tabular} \\
    \midrule
    BC-RNN & $0.0\pm0.0$ & $0.0\pm0.0$ & $48.7\pm18.6$ \\
    HBC & $\mathbf{3.3\pm2.1}$ & $9.3\pm5.3$ & $71.3\pm6.5$ \\
    d-HBC & $1.1 \pm 1.7$ & $4.4 \pm 4.9$ & $33.3\pm9.7$ \\
    r-HBC (ours) & $\mathbf{3.3\pm2.1}$ & $\mathbf{17.3\pm5.3}$ & $\mathbf{86.7\pm5.2}$ \\
    \bottomrule
  \end{tabular}
\vspace{-15pt}
\end{table}
\endgroup

Since the coordinated subtasks require implicit communication between operators and are more subject to system issues such as latency, we expect coordination to be the major bottleneck of collecting successful demonstrations. To quantify if coordination was an issue, we examine the difficulty of our tasks by evaluating the marginal degradation that each type of sub-task contributes to operator task completion rate. 
For the Assembly task and Pick-Place Handover task, both tasks first have an uncoordinated subtask followed by a coordinated subtask. We therefore measure the marginal degradation of the uncoordinated subtask by measuring the difference between its best possible success rate (100\%) and the uncoordinated subtask success rate. The degradation is measured for the coordinated subtask by calculating the difference between its best possible success rate (i.e. the uncoordinated subtask success rate) and the coordinated subtask success rate. For the Lift Wipe task, since the order of the subtasks is reversed with coordinated subtask being followed by the uncoordinated subtask, we reverse the order of calculations.

Table~\ref{table:coordination} demonstrates that for the two-arm tasks, the marginal degradation of uncoordinated subtasks were higher than for coordinated subtasks by roughly $20\%$, meaning that operators failed more frequently on the uncoordinated subtask sections. For the three-arm task we see that the degradation rate for the coordinated subtask is slightly higher ($9\%$). Taken together, these results show that coordination does not pose a significant barrier to operators for completing a task demonstration successfully, highlighting that \sysName is suitable for collecting collaborative task demonstrations despite operators being physically separated by large distances.

\subsection{Data Analysis}
\label{sec:results}

We evaluate all models on experienced-user demonstrations collected for all tasks, seen in Table~\ref{task_benchmark_algo_results}. We also evaluate a subset of models on demonstrations collected during our user study, presented in Table~\ref{task_user_study_algo_results}. We record the best checkpoint rollout success rate over the course of training, and report the mean and standard deviation across five random seeds.


\textbf{Are centralized and decentralized variants of standard IL methods sufficient for learning from multi-arm task demonstrations?} We first discuss our two single-stage tasks. d-HBC outperforms HBC by a wide margin ($84.5\%$ vs. $38.5\%$) on the Multi-Cube Lifting task. This is expected since human operators lifted their own cubes independently. Interestingly, d-HBC and HBC perform comparably on the Drink-Tray Lifting task. We hypothesize that this is because the task is short-horizon and the demonstrators grasped each handle at roughly the same time, allowing each independent agent in d-HBC to just focus on grasping its handle and lifting independent of the other agent. Indeed, on the longer horizon Three Arm Lifting Wiping task, where the arms must coordinate to lift and move the tray for longer periods of time, we see HBC outperforms d-HBC ($83.7\%$ vs. $50.0\%$).

On the Handover task, d-HBC slightly outperforms HBC ($24.4\%$ vs. $16.0\%$). This might be because significant portions of the Handover task do not require the arms to be aware of each other's actions. On the Assembly task, both perform poorly ($\sim 5\%$). Based on these results, we conclude that for our more challenging multi-stage tasks, neither d-HBC nor HBC consistently outperforms the other. We also note that the BC-RNN baseline performs poorly across all tasks compared to HBC and the other variants, highlighting the substantial benefits of hierarchy in the multi-arm setting.


\textbf{Can partially decentralized hierarchical models sufficiently capture mixed coordination properties to better succeed at multi-arm tasks?} Our naive variations dh-HBC and dl-HBC at best perform marginally better than the lowest performing centralized or decentralized HBC baseline, and sometimes perform worse than both baselines, as in the Drink-Tray Lifting ($<70\%$) and Pick-Place Handover ($<16\%$) tasks. These results highlight how mixed coordinated settings cannot easily be solved with naive approaches.


\textbf{Can our proposed residual framework better capture mixed coordination properties to improve policy performance on multi-arm tasks?} In contrast to the partially decentralized baselines, our residual models r-HBC and rd-HBC consistently outperform all baselines across all of our tasks. We hypothesize that because our residual model allows for small action perturbations, our framework can produce a policy that endows the base policy with complementary behavior in states that incur high action error, without compromising base policy behavior in well-fit states.

The consistent performance improvements exhibited by our residual-augmented policies highlight the potential of our framework to be applied to a wide range of multi-arm tasks with varying levels of mixed coordination, from the highly coordinated instance (Three Arm Lifting Wiping) to the weakly coordinated case (Two Arm Pick-Place Handover). We also observed that rd-HBC performed best in the short-horizon tasks such as Drink-Tray Lifting ($86.7\%$ vs. $75.3\%$), whereas r-HBC outperformed in the more complex, multi-stage tasks such as Lifting Wiping ($94.0\%$ vs. $58.6\%$), highlighting how inductive bias still plays a major role in choosing a suitable base policy that may lead to the best success rates. 

\textbf{How robust is our proposed residual framework to varying demonstration quality?} We expect model performance to degrade as demonstration quality reduces due to less-experienced operators, and find that our r-HBC model still performs as well or better ($17.3\%$ vs. $9.3\%$ for Pick-Place Handover, $86.7\%$ vs. $71.3\%$ for Lifting Wiping) than our other baselines in that condition. This shows that our proposed model is robust enough to improve performance despite noisy training signals, and can learn from a diverse distribution of demonstrations.

\textbf{What are the limitations of the proposed residual framework?}
While our residual framework has shown promising results in improving current multi-arm IL methods for multi-arm tasks, we observe room to improve, especially in the more challenging tasks such as the Assembly and Pick-Place Handover tasks. 
While we defer this to future work, we highlight \sysName as the means for conveniently gathering data necessary to explore these novel emergent properties underlying such multi-arm tasks.

\section{Conclusion}
We introduced \sysName, a scalable teleoperation system for gathering real-time multi-arm manipulation task demonstrations, and showed that IL methods can leverage this data to train performant policies over a wide range of realistic and novel multi-arm tasks requiring varying degrees of collaboration. We also explored potential methods for better modeling mixed coordination policies, and showed that a residual-augmented framework is able to outperform all of our other baselines on our tasks. Imitation learning for multi-arm manipulation has been limited due to the difficulty of collecting demonstrations, but we are excited by the prospect of \sysName lowering this barrier and enabling further research in this setting.
\label{sec:conclusion}

\clearpage

{\footnotesize 
\section*{Acknowledgment}
We would like to thank Rohun Kulkarni and Margaret Tung for helping with data collection. Ajay Mandlekar acknowledges the support of the Department of Defense (DoD) through the NDSEG program. We acknowledge the support of Toyota Research Institute (``TRI''); this article solely reflects the
opinions and conclusions of its authors and not TRI or any other Toyota entity.
}

\renewcommand*{\bibfont}{\footnotesize}
\printbibliography 

\end{document}